\documentclass[runningheads]{llncs}

\usepackage{eccv}

\usepackage{eccvabbrv}

\usepackage{graphicx}
\usepackage{booktabs}

\usepackage[accsupp]{axessibility}  
\usepackage{booktabs}      
\usepackage[table]{xcolor} 
\usepackage{graphicx}      
\usepackage{array}
\usepackage{booktabs}
\usepackage{colortbl}
\usepackage{multirow}
\usepackage{amsfonts}
\usepackage{nicematrix} 
\usepackage{tabularx}

\usepackage{hyperref}

\usepackage{orcidlink}

\usepackage{multirow}
\begin{document}

\title{SpatialBench: Benchmarking Multimodal Large Language Models for Spatial Cognition} 

\titlerunning{Abbreviated paper title}

\author{Peiran Xu\inst{1,7} \and
Sudong Wang\inst{2} \and
Yao Zhu\inst{3} \and
Jianing Li\inst{4} \and
Gege Qi\inst{5} \and 
Yunjian Zhang\inst{6}
}

\authorrunning{P.Xu et al.}

\institute{
Sun Yat-Sen University \and
HKUST (GZ) \and
Zhejiang University \and
Peking University \and
CAICT \and
UCAS \and
CUC
}

\maketitle

\begin{abstract}
  Spatial cognition is fundamental to real-world multimodal intelligence, allowing models to effectively interact with the physical environment. While multimodal large language models (MLLMs) have made significant strides, existing benchmarks often oversimplify spatial cognition, reducing it to a single-dimensional metric, which fails to capture the hierarchical structure and interdependence of spatial abilities. To address this gap, we propose a hierarchical spatial cognition framework that decomposes spatial intelligence into five progressively complex levels from basic observation to high-level planning. Building upon this taxonomy, we construct SpatialBench, a large-scale, fine-grained benchmark covering 15 tasks aligned with these cognitive levels. To provide a unified evaluation across heterogeneous tasks, we further introduce a high-level capability–oriented metric that reliably assesses a model’s overall spatial reasoning ability. Extensive experiments over massive MLLMs reveal distinct performance stratification across cognitive levels: models exhibit strong perceptual grounding yet remain limited in symbolic reasoning, causal inference, and planning. Additional human tests demonstrate that humans perform selective, goal-directed abstraction, while MLLMs tend to over-attend to surface details without coherent spatial intent. Our work establishes the first systematic framework for measuring hierarchical spatial cognition in MLLMs, laying the foundation for future spatially intelligent systems.
  \keywords{MLLMs \and Spatial Cognition \and Benchmark}
\end{abstract}

\section{Introduction}
\label{sec:intro}
In daily life, human can effortlessly integrate spatial information from their surroundings, with a capability known as spatial cognition. This ability extends beyond mere object recognition, serving as a cognitive bridge between perceptual inputs and higher-level functions such as reasoning and navigation. 
With the rapid advancements of large language models (LLMs) \cite{brown2020language,vicuna2023,driess2023palm,gilardi2023chatgpt,rohan2023alpaca,touvron2023llama}, multimodal large language models (MLLMs) have recently emerged as a major step toward general-purpose visual–linguistic intelligence \cite{bai2023qwen,dai2023instructblip,li2023blip,zhang2023internlm,brooks2023instructpix2pix,black2023training,li2023llavamed,zhu2023minigpt,zhang2023gpt4roi,liu2023llava,liu2023improvedllava,ye2023mplug,He2024malmm,Zhang2024groundhog,Chen2023internvl,Yuan2023osprey,Dong2024dreamllm,Cha2023honeybee,qwq32b}. By jointly aligning visual and textual modalities within a shared semantic space, MLLMs have moved beyond abstract visual representations, integrating linguistic context to interpret scenes in a more structured manner. 
Recent advances show that MLLMs have exhibited spatial reasoning abilities \cite{Cai2025spatial,Cheng2024spatial,Chen2024spatial,Han2025video,Li2024top,Yamada2024evaluating,zhu2024llava,kumar2025does,yang2023set,tang2024sparkle,wu2025spatial,li2025llava,liu2025oryx}, and several benchmarks have been introduced to quantify these capabilities \cite{azuma2022scanqa,ma2023sqa,yang2025thinking,li2025spatial,yin2025spatial,tong2024cambrian,zhang2025from,li2025view}. However, they remain fragmented and task-oriented, often emphasizing performance on specific vision–language tasks rather than assessing spatial cognition as a structured capability. In addition, most benchmarks rely on synthetic or narrowly defined datasets, lacking the visual diversity and real-world complexity necessary to probe genuine spatial cognition. Consequently, these evaluations provide only a partial view of spatial intelligence, making it difficult to analyze cognitive processes and reveal systematic deficiencies in models.

\begin{figure}
    \centering
    \includegraphics[width=.45\linewidth]{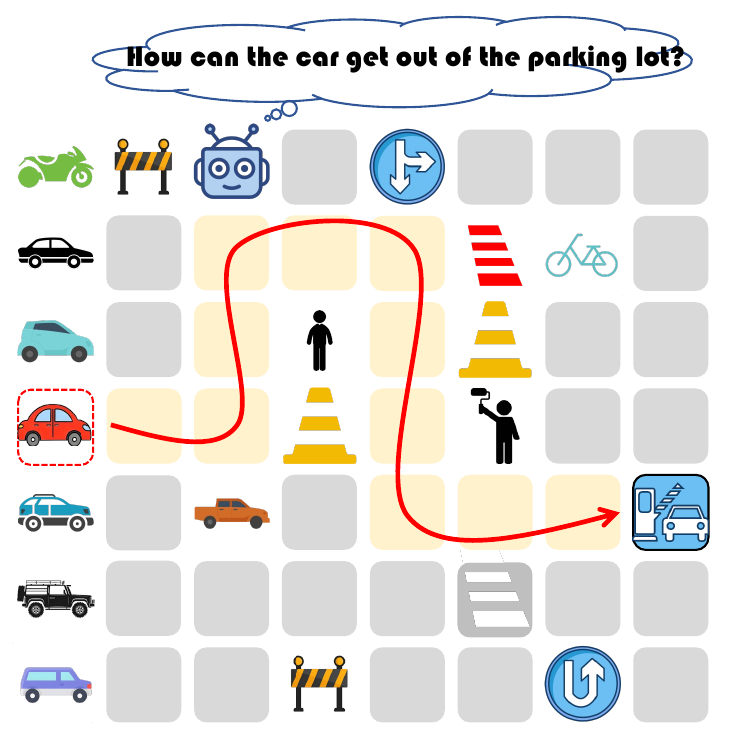}
    \caption{In a parking lot, the vehicle must understand the relationships of surrounding objects, reason about possible events, and plan the optimal route to reach the exit.}
    \label{fig:parking}
\end{figure}
\vspace{-0.3cm}
To overcome these limitations, we propose a cognitively grounded evaluation framework for spatial intelligence in MLLMs. Inspired by the cognitive map theory \cite{tolman1948cognitive,o1979precis,battaglia2013simulation} in neuroscience, we conceptualize spatial cognition as a hierarchical process that evolves from low-level perception to high-level reasoning and decision making. Specifically, our framework decomposes spatial understanding into five progressive levels, including \textbf{observation (L1)}, \textbf{topology and relation (L2)}, \textbf{symbolic reasoning (L3)}, \textbf{causality (L4)}, and \textbf{planning (L5)}, each corresponding to distinct cognitive functions involved in human intelligence. For the example in Fig. \ref{fig:parking}, consider a scenario where a car leaves a parking lot. The model recognizes relevant entities and their spatial configurations (L1), then it understands topological relations such as lane connectivity and obstructions (L2) and maps visual symbols to semantic meanings and evaluates potential detour options (L3). After that, it infers causal outcomes of possible maneuvers (L4), and finally, it integrates prior reasoning to generate a coherent plan (L5). This hierarchical design provides a structured lens through which to interpret model behavior, enabling ability-oriented rather than task-oriented evaluation of spatial intelligence.

Building upon this framework, we construct a large-scale spatial video dataset named SpatialBench that grounds spatial cognition evaluation in realistic scenarios. Unlike previous synthetic or narrowly scoped datasets, our collection is captured from diverse indoor and outdoor environments, encompassing both static spatial layouts and dynamic scene evolutions that reflect the multi-level cognitive demands of spatial intelligence. To realize the five-level cognitive hierarchy, we design 15 categories of spatial reasoning tasks, each aligned with a distinct stage of spatial cognition. Each video is paired with carefully designed questions and annotations aligned with these cognitive dimensions, enabling systematic, fine-grained, and cognitively interpretable assessment for MLLMs.

Our experiments show that although current MLLMs perform well on perceptual and relational reasoning tasks, their competence declines sharply in high-level tasks. Insights from the one-shot and human benchmarks suggest that humans rely on selective, goal-oriented reasoning, while MLLMs exhibit diffuse attention to scene details, lacking a unified spatial cognition. 

Our contributions are summarized as follows:
\begin{itemize}
    \item We establish the first comprehensive and cognitively grounded framework for assessing spatial intelligence of MLLMs. Drawing inspiration from cognitive map theory, our framework hierarchically decomposes spatial cognition into five progressive levels, shifting evaluation from task-driven to ability-oriented assessment.
    \item We construct SpatialBench, a large-scale multimodal dataset specifically designed for evaluating spatial cognition in MLLMs. It features 15 distinct categories of spatial reasoning tasks aligned with five hierarchical cognitive levels, providing a robust foundation for systematic and scalable evaluation.
    \item We introduce a high-level ability-driven evaluation metric to assess spatial cognition in MLLMs. Through extensive experiments on a wide range of state-of-the-art open-source and commercial models, we uncover their strengths and limitations in spatial reasoning. We further conduct controlled human evaluations to compare human and model reasoning, offering new insights into the gap between artificial and human spatial intelligence.
\end{itemize}
\section{Related Works}
\label{sec:rel}
\subsection{Multimodal Large Language Models}
Recent progress in Large Language Models (LLMs) \cite{brown2020language,vicuna2023,driess2023palm,gilardi2023chatgpt,rohan2023alpaca,touvron2023llama} has catalyzed the evolution of MLLMs, which integrate visual and linguistic modalities within a unified semantic space. By aligning image representations with textual instructions, MLLMs demonstrate remarkable capability in understanding and generating multimodal content across a wide spectrum of real-world tasks \cite{chen2023sharegpt4v,huang2023diversity,alayrac2022flamingo,chen2022pali,driess2023palm}. In general, an MLLM consists of three core components: a modality encoder \cite{radford2021learning,li2022blip}, a language backbone (LLM), and a modality interface that connects the two. 
Recently, the capability of MLLMs has expanded beyond static images to encompass video understanding \cite{chen2024ll3da,fu2024scene,qi2025gpt4scene}. This advancement has led to the incorporation of video–language alignment during pre-training, allowing MLLMs to jointly model temporal semantics and motion dynamics within a unified multimodal framework \cite{lin2024video}.

\subsection{Visual-based Spatial Cognition}
Visual-based spatial cognition seeks to endow MLLMs with the ability to perceive and reason about three-dimensional spatial relationships directly from visual inputs \cite{Cheng2024spatial,Chen2024spatial,Han2025video,Li2024top,Yamada2024evaluating,mangalam2023egoschema}. Several benchmarks have emerged to evaluate this capability from different perspectives. VSP \cite{wu2024vsp} highlights the bottleneck of perception and reasoning in spatial planning, focusing on static scenes with picture inputs. Video-MME \cite{fu2025videomme} provides a comprehensive assessment across a range of video-related tasks involving recognition and perception. VLM4D \cite{zhou2025vlm4d} emphasizes dynamic motion analysis, and STI-Bench \cite{li2025sti} examines physical reasoning by testing models’ ability to predict and estimate object motions and displacements. VSI-Bench \cite{yang2025thinking} introduces a structured benchmark comprising eight question types, each designed to probe different dimensions of spatial understanding in MLLMs. Ego-ST Bench \cite{wu2025st} extends this to self-centered navigation in egocentric environments. SpatialLadder \cite{li2025spatial} presents a comprehensive dataset spanning multiple categories, covering spatial reasoning tasks from single-image understanding to video-based inference. MindCube \cite{yin2025spatial} assesses MLLMs’ ability to infer complete spatial structures from limited visual observations.
\vspace{-0.3cm}
\begin{table*}[]
\centering
\caption{Comparison of \textbf{SpatialBench} with existing spatial and video benchmarks.}
\label{tab:benchmark_comparison}
\resizebox{\textwidth}{!}{
\begin{tabular}{lccccc}
\toprule
\textbf{Benchmark} & \textbf{Data Nature} & \textbf{Modality} & \textbf{Perspectives} & \textbf{Taxonomy} & \textbf{Scale (Vid/QA)} \\ \midrule
VSP & Synthetic \& 2D images & Image & - & 4 Levels & - / 4.4k \\
VLM4D & Hybrid & Video & Ego \& Exo & Flat & 1k / 1.8k \\
Video-MME & Mixed & Video & Mixed & Flat & 900 / 2.7k \\
VSI-Bench & Open Source Datasets (Indoor) & Video & Ego & 3 Levels & 288 / 5k+ \\
STI-Bench & Real & Video & Ego \& Exo & 2 Levels & 300 / 2k \\ \midrule
\rowcolor[HTML]{F2F2F2} 
\textbf{SpatialBench (Ours)} & \textbf{Self-recorded (In/Outdoor)} & \textbf{Video} & \textbf{Ego \& Exo} & \textbf{5 Levels} & \textbf{117 / 3.1k+} \\ \bottomrule
\end{tabular}
}
\end{table*}

While simulation-based benchmarks like VSP offer controlled environments, they often fail to capture the physical nuances of the real world. Other real-life video based benchmarks like VSI-Bench utilize orderly and open-source indoor videos, lacking noise interference in practical application scenarios (such as outdoor robots). SpatialBench distinguishes itself by featuring diverse real-world video sequences across both indoor and outdoor settings, challenging MLLMs with complex lighting, sensor noise, and authentic 3D geometry that are absent in synthetic data. We present the differences between SpatialBench and other benchmarks in Table \ref{tab:benchmark_comparison}.

\section{Spatial Cognition Ability Framework}
\label{sec:spatial}
\subsection{Conceptual Foundations of Spatial Cognition}
Spatial cognition encompasses the processes that enable intelligent systems to perceive, represent, and reason about spatial relationships within their environment. It involves acquiring spatial knowledge from sensory inputs, forming internal representations of space, and utilizing these representations for high-level tasks \cite{tolman1948cognitive,kozhevnikov2023different}. The cognitive map theory \cite{tolman1948cognitive,o1979precis} provides a foundational opinion of how such representations are organized. Originating from behavioral studies of animals and later supported by discoveries of place cells in the hippocampal system, this theory proposes that intelligent agents construct internal, map-like structures that encode both metric and topological relations. These cognitive maps allow flexible navigation, route planning, and generalization beyond direct sensory experience. More importantly, they reveal that spatial knowledge is not a flat or static representation, but a hierarchically organized system, where low-level perceptual and motor cues are progressively abstracted into higher-order representations that integrate semantic, relational, and causal information. Building on this foundation, modern computational perspectives extend spatial cognition toward causal and multimodal understanding \cite{battaglia2013simulation,lake2017building,botvinick2017building}.
This hierarchical and integrative view provides the theoretical basis for our framework, which conceptualizes spatial cognition in MLLMs as a progressive process evolving from perception to reasoning and ultimately to planning.

\begin{figure}
    \centering
    \includegraphics[width=.6\linewidth]{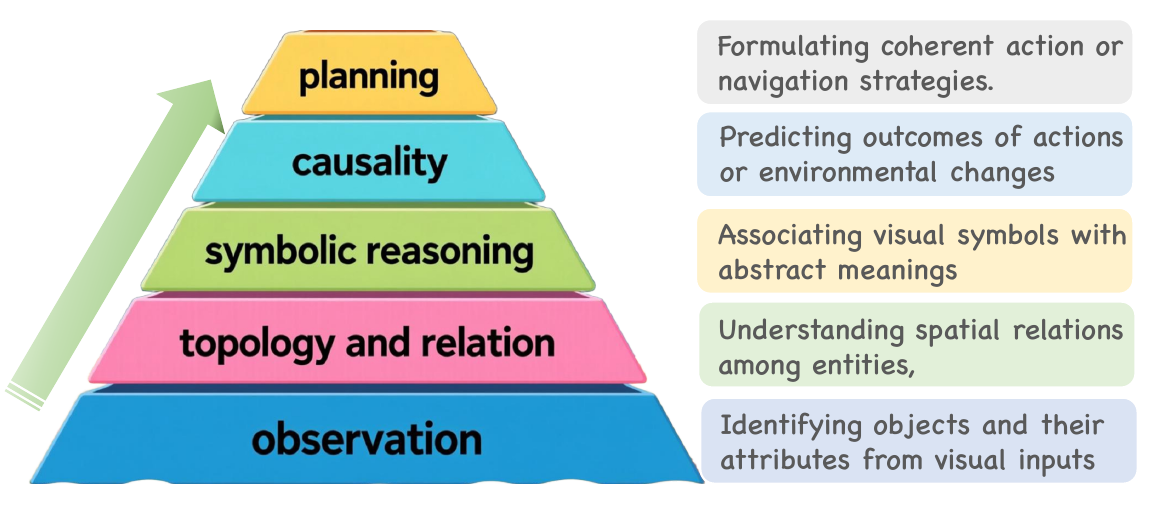}
    \caption{The proposed hierarchical spatial cognitive taxonomy.}
    \label{fig:cognition}
\end{figure}
\vspace{-0.9cm}
\subsection{Hierarchical Spatial Cognitive Taxonomy}
Based on the cognitive map theory, we propose the first systematic and hierarchical framework for spatial cognition evaluation, capturing the progressive development of spatial understanding from perception to high-level reasoning. Unlike prior benchmarks that focus on isolated visual tasks, our taxonomy is ability-driven, where each level represents a distinct, measurable cognitive capacity that reflects a specific stage of spatial intelligence. This hierarchical framework delineates spatial cognition into five progressive levels, each corresponding to a fundamental stage in the transition from sensory perception to deliberative reasoning: \textbf{observation (L1)}, \textbf{topology and relation(L2)}, \textbf{symbolic reasoning (L3)}, \textbf{causality (L4)}, and \textbf{planning (L5)}. Together, these levels illustrate the progressive process by which intelligent systems transform raw perceptual information into organized spatial reasoning, as shown in Fig. \ref{fig:cognition}.

\noindent\textbf{Observation.} The foundational level of spatial cognition is observation, where the model identifies objects and their attributes from visual inputs. This stage corresponds to the extraction of basic perceptual elements such as object category, color, shape, and size.

\noindent\textbf{Topology and relation.} This level focuses on spatial relations among entities, such as adjacency, containment, orientation, and connectivity. Rather than perceiving isolated objects, it concerns the relational structure of the environment, describing how different elements are spatially arranged and interact within a coherent scene configuration.

\noindent\textbf{Symbolic reasoning.} Spatial understanding is extended beyond geometry into semantic interpretation. The agent is expected to associate visual symbols or spatial cues (e.g., arrows, pathways) with their abstract meanings and apply rule-based reasoning to infer spatial intent or constraints.

\noindent\textbf{Causality.} This level reflects the ability to infer spatiotemporal dependencies and predict outcomes of actions. It involves reasoning about how object movements, physical interactions, or agent behaviors lead to specific consequences, integrating physical intuition and causal understanding into reasoning.

\noindent\textbf{Planning.} It represents the highest stage of spatial cognition, where perception, relational understanding, and causal reasoning are integrated to enable deliberate, goal-oriented decision making. At this level, an agent should synthesize its spatial representations and predictive reasoning to formulate coherent action sequences or navigation strategies that adapt to dynamic environmental contexts.

Collectively, these five levels delineate a progressive and interdependent hierarchy, where foundational sensory capabilities serve as the indispensable bedrock for higher-order reasoning and planning. This structural dependency marks a significant departure from existing benchmarks, which predominantly evaluate model efficacy through the narrow lens of isolated, fragmented tasks. Rather than yielding disjointed performance metrics, our taxonomy enables researchers to pinpoint the precise developmental stage of a model's spatial cognition and diagnose its cognitive bottlenecks. By organizing spatial understanding into this ability-driven, cumulative continuum, we provide the first systematic and cognitively grounded paradigm. It offers a unified foundation not merely for benchmarking, but for interpreting the evolutionary trajectory of spatial intelligence across MLLMs of varying architectures and scales.
\begin{figure}
    \centering
    \includegraphics[width=1.\linewidth]{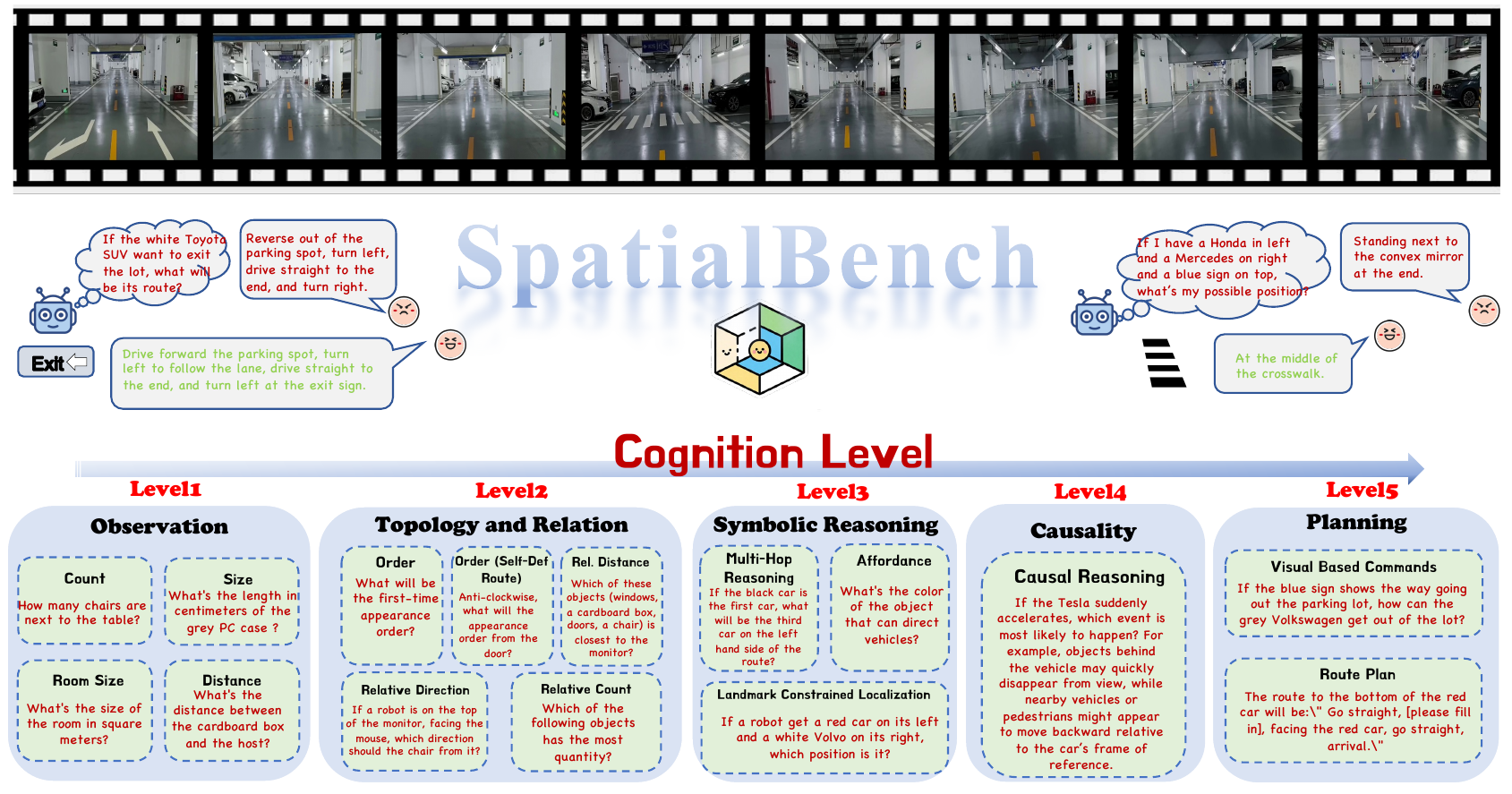}
    \caption{An overview of SpatialBench.}
    \label{fig:framework}
\end{figure}

\section{SpatialBench}
\label{sec:bench}
\subsection{Overview}
We introduce SpatialBench, as shown in Fig. \ref{fig:framework}, a large-scale benchmark for assessing evaluate the hierarchical spatial cognition of MLLMs using first-person videos. The dataset comprises 15 question types, each carefully mapped to one of the five cognitive levels introduced above:
\begin{itemize}
    \item Observation (L1): object counting, object size, room size, absolute distance;
    \item Topology and relation (L2): appearance order, relative distance, relative direction, appearance order on self-defined route, relative counting;
    \item Symbolic reasoning (L3): multi-hop spatial reasoning, affordance, landmark-constrained pose localization;
    \item Causality (L4): spatial causal reasoning;
    \item Planning (L5): visual-based commands, route planning.
\end{itemize}

\begin{figure}
    \centering
    \includegraphics[width=.7\linewidth]{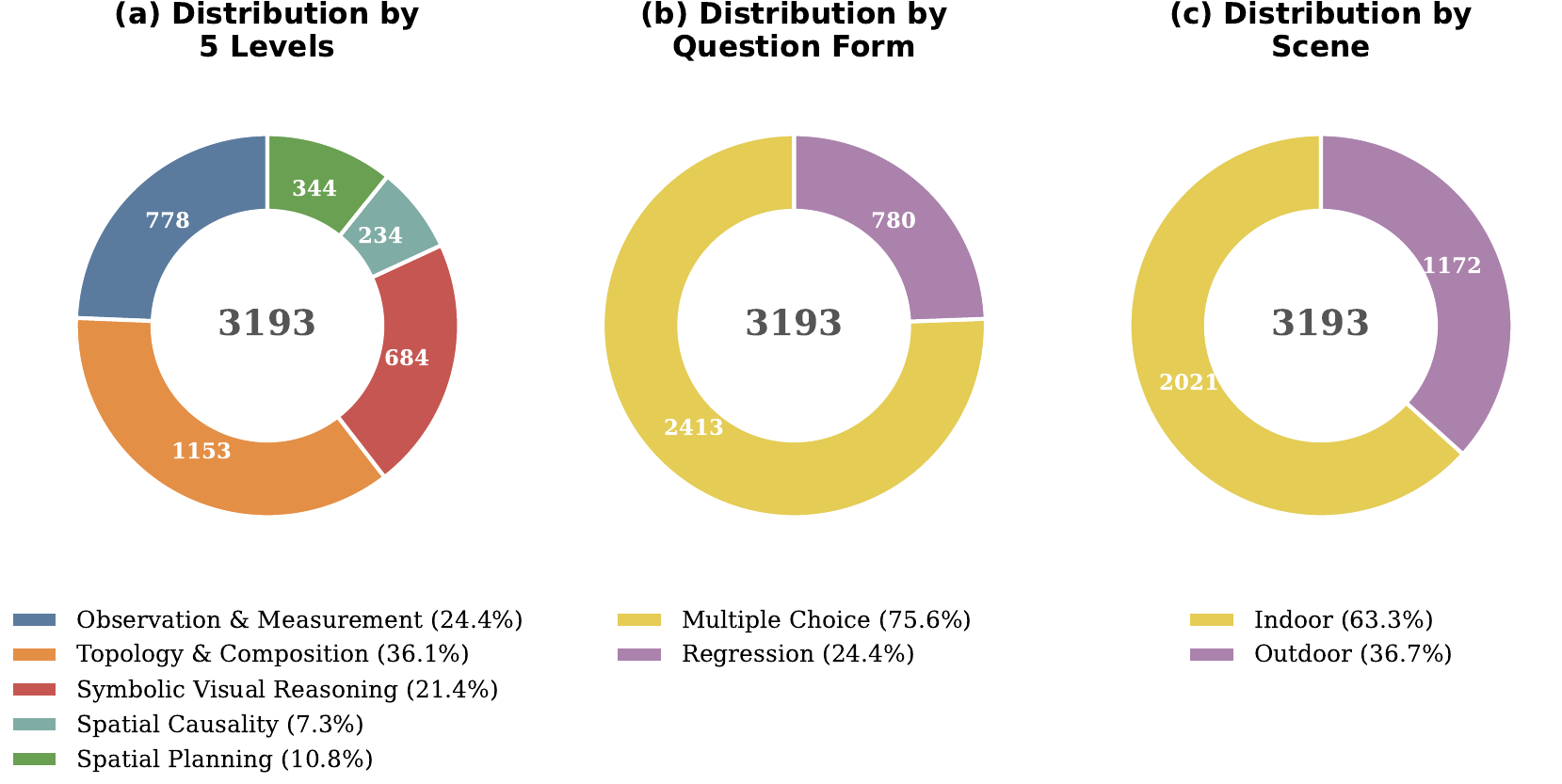}
    \caption{The statistics of SpatialBench.}
    \label{fig:statistics}
\end{figure}

The SpatialBench dataset comprises 3193 question–answer pairs sourced from 117 videos captured by us in real life. The collection spans both indoor and outdoor settings and includes static as well as dynamic scenes, covering challenging real-world contexts such as city roads, forest trails, residential areas, and underground environments. Together, these recordings provide diverse temporal and spatial complexity for evaluating MLLMs’ spatial cognition ability in realistic scenarios. The detailed statistics of SpatialBench are shown in Fig. \ref{fig:statistics}.

\subsection{Benchmark Construction}
\noindent\textbf{Data Collection.} In contrast to prior benchmarks that adapt existing open-source datasets, SpatialBench is built from scratch through real-world recordings using our custom-designed sensing platform. The platform integrates a calibrated RGB camera and a 3D LiDAR sensor, which are spatially and temporally synchronized to ensure precise correspondence between visual and geometric modalities. The RGB camera continuously captures high-resolution visual streams, which serves as the basis for video question generation, while the LiDAR sensor synchronously records 3D point clouds that provide precise geometric information for size and distance related measurements. To ensure the accuracy and density of spatial data, we apply a filtering procedure to remove overly sparse or noisy point clouds. Data are collected from diverse environments, covering both indoor and outdoor settings such as offices, residential areas, city streets, and wooded regions, including both dynamic and static scenes. Each recording is conducted from a first-person perspective to preserve the egocentric characteristics essential for spatial understanding. For every video, we record standardized metadata including timestamps, scene categories, LiDAR frames, and synchronization parameters.

\noindent\textbf{Question-Answer Generation.} To ensure high-quality and semantically diverse annotations, human annotators work in pairs throughout the QA generation process. Within each pair, one annotator proposes candidate questions while the other independently reviews and validates them, checking for duplicates, ambiguous wording, and alignment with the intended cognitive level. All annotators are trained on the fifteen predefined task types, and they carefully review each video segment and propose candidate questions grounded in observed spatial relationships and scene dynamics. This collaborative review process ensures both the accuracy and relevance of the human-generated questions before they proceed to AI-assisted answer generation. For non-metric question types, we design specialized prompting templates tailored to each question category. These templates are then provided to state-of-the-art commercial models to generate corresponding answers. Along with each generated response, the model is required to output an evidence summary, including key frames and brief reasoning traces. For metric-related questions, we directly compute ground-truth answers using the LiDAR point cloud data. Precise 3D measurements are extracted through geometric fitting and spatial projection, thereby providing physically accurate ground-truth answers for all size and distance related questions. This hybrid design combines the semantic richness of human understanding, the efficiency of large model reasoning, and the geometric accuracy of sensor-derived measurements, resulting in a balanced and cognitively interpretable QA corpus.

\noindent\textbf{Annotation Verification.} We implement a multi-step verification protocol to ensure the reliability of the generated annotations. For L1 and L2 questions, multiple leading models independently generate answers, and the consistency of these responses is evaluated. Questions with fully consistent model outputs are provisionally approved, but a subset of these automatically approved answers is further subject to human spot-checking to guarantee overall quality. Any discrepancies detected, whether during consistency checks or spot audits, trigger full human review. For question beyond L3, all annotations undergo mandatory human verification due to their higher cognitive complexity. The human review process follows a fixed checklist including: whether the evidence frames display key entities, whether multi-model outputs have been correctly interpreted, and whether answers conform to the predefined question schema. Annotations that are modified or rejected during the first review are subsequently evaluated by an additional annotator. This annotator examines the evidence summaries, the explanations and supporting screenshots submitted by the human reviewers, and also inspects the video and the question directly. Based on this comprehensive review, this annotator determines the final annotation. The verification protocol used ensures that all QA annotations are accurate, consistent, auditable, and reproducible, providing a trustworthy foundation for evaluation.
\begin{table*}[t]
\centering
\caption{Comprehensive evaluation results on SpatialBench. Models are grouped by category and sorted by their overall Rank. \textbf{Bold} indicates the best performance overall; \underline{underline} indicates the best performance among open-source models. For presentation, we simplified the table.}
\label{tab:result-com}
\small
\resizebox{\textwidth}{!}{
\begin{tabular}{l|cc|ccccc}
\toprule
\textbf{Model} & \textbf{Score (Rank)} & \textbf{Ave.} & \textbf{Obs.} & \textbf{Topo.} & \textbf{Symb.} & \textbf{Caus.} & \textbf{Plan.} \\
\hline
Human Level & 96.40 (-) & 93.89 & 77.78 & 100.00 & 100.00 & 100.00 & 100.00 \\
Random & 25.32 (-) & 25.06 & - & 24.97 & 24.95 & 25.57 & 25.50 \\
\hline
\rowcolor{gray!10}\textit{Proprietary Models} & & & & & & & \\
Gemini-2.5-pro & \textbf{75.79 (1)} & \textbf{71.86} & \textbf{55.63} & \textbf{72.82} & \textbf{84.88} & \textbf{80.81} & \textbf{74.11} \\
GPT-4o-mini & 30.92 (9) & 33.44 & 36.12 & 40.73 & 21.65 & 33.33 & 27.68 \\
Claude-sonnet-4-5 & 28.56 (11) & 32.10 & 50.27 & 24.38 & 30.96 & 22.50 & 27.78 \\
GPT-5-chat-latest & 22.45 (14) & 23.93 & 48.48 & 15.01 & 15.83 & 20.00 & 24.07 \\
\hline
\rowcolor{gray!10}\textit{Open Source Models} & & & & & & & \\
Cambrian-S-3B & \underline{59.12 (2)} & \underline{50.90} & 31.67 & 48.43 & \underline{65.34} & 64.22 & \underline{65.00} \\
VST-3B-RL & 58.29 (3) & 48.35 & 22.71 & \underline{49.87} & 60.62 & \underline{69.83} & 62.35 \\
Qwen3-VL-235B-A22B-Inst & 37.79 (4) & 38.84 & 48.68 & 37.40 & 32.99 & 33.33 & 41.07 \\
Qwen3-VL-235B-A22B-Thin & 36.27 (5) & 36.60 & 47.70 & 33.33 & 29.55 & 28.28 & 44.64 \\
Qwen2.5-VL-72B-Inst & 33.82 (6) & 32.73 & 38.33 & 29.01 & 33.33 & 36.36 & 33.04 \\
Qwen3-VL-30B-A3B-Inst & 32.83 (7) & 37.07 & 48.52 & 41.78 & 21.31 & 29.29 & 33.04 \\
Qwen3-VL-30B-A3B-Thin & 31.22 (8) & 36.83 & 46.20 & 41.46 & 26.12 & 26.26 & 28.57 \\
GLM-4V-Plus-0111 & 30.40 (10) & 36.89 & 49.81 & 40.90 & 23.71 & 36.73 & 17.86 \\
ERNIE-4.5-Turbo-VL-32K & 28.05 (12) & 34.72 & 44.25 & 41.58 & 20.62 & 30.30 & 18.75 \\
GLM-4.5V & 27.72 (13) & 35.34 & \underline{50.16} & 38.21 & 25.00 & 28.12 & 16.96 \\
Qwen2.5-VL-7B-Instruct & 17.45 (15) & 17.84 & 19.69 & 15.48 & 17.96 & 18.95 & 16.22 \\
\bottomrule
\end{tabular}
}
\end{table*}

\begin{table*}
    \caption{Detailed evaluation results on SpatialBench for 15 tasks.}
    \label{tab:result}
    \resizebox{\textwidth}{!}{
    \begin{tabular}{r|cccc|ccccc|ccc|c|cc}
    \toprule
    \renewcommand{\arraystretch}{1.25}
    & 
    \textbf{\cellcolor{yellow!20}{\rotatebox{75}{Obj.count}}}&
    \textbf{\cellcolor{yellow!20}{\rotatebox{75}{Obj.Size}}}&
    \textbf{\cellcolor{yellow!20}{\rotatebox{75}{Room Size}}}&
    \textbf{\cellcolor{yellow!20}{\rotatebox{75}{Abs. Distance}}}&
    \textbf{\cellcolor{orange!20}{\rotatebox{75}{App. Order}}}&
    \textbf{\cellcolor{orange!20}{\rotatebox{75}{\shortstack{App. Order\\(Self-Def\\Route)}}}}&
    \textbf{\cellcolor{orange!20}{\rotatebox{75}{Rel. Distance}}}&
    \textbf{\cellcolor{orange!20}{\rotatebox{75}{Rel. Direction}}}&
    \textbf{\cellcolor{orange!20}{\rotatebox{75}{Rel. Count}}}&
    \textbf{\cellcolor{SkyBlue!40}{\rotatebox{75}{\shortstack{Multi-Hop\\Reasoning}}}}&
    \textbf{\cellcolor{SkyBlue!40}{\rotatebox{75}{Affordance}}}&
    \textbf{\cellcolor{SkyBlue!40}{\rotatebox{75}{\shortstack{Landmark\\Constrained\\Location}}}}&
    \textbf{\cellcolor{SkyBlue!20}{\rotatebox{75}{\shortstack{Causal\\Reasoning}}}}&
    \textbf{\cellcolor{green!20}{\rotatebox{75}{\shortstack{Visual Based\\Commands}}}}&
    \textbf{\cellcolor{green!20}{\rotatebox{75}{Route Plan}}}\\
    \hline
    \textbf{Model}&
    \multicolumn{4}{|c|}{\textbf{\cellcolor{yellow!20}Observation}}&
    \multicolumn{5}{|c|}{\textbf{\cellcolor{orange!20}Topology \& Relation}}&
    \multicolumn{3}{|c|}{\textbf{\cellcolor{SkyBlue!40}Symbolic Reasoning}}&
    \multicolumn{1}{|c|}{\textbf{\cellcolor{SkyBlue!20}Causality}}&
    \multicolumn{2}{|c}{\textbf{\cellcolor{green!20}Planning}}\\
    \hline
    \textbf{Human Level}&\cellcolor{yellow!20}{100.00}&\cellcolor{yellow!20}{53.75}&\cellcolor{yellow!20}{67.50}&\cellcolor{yellow!20}{60.00}&\cellcolor{orange!20}{100.00}&\cellcolor{orange!20}{100.00}&\cellcolor{orange!20}{100.00}&\cellcolor{orange!20}{100.00}&\cellcolor{orange!20}{100.00}&\cellcolor{SkyBlue!40}100.00&\cellcolor{SkyBlue!40}100.00&\cellcolor{SkyBlue!40}100.00&\cellcolor{SkyBlue!20}100.00&\cellcolor{green!20}100.00&\cellcolor{green!20}100.00\\
    \hline
    \textbf{Random}&\cellcolor{yellow!20}{-}&\cellcolor{yellow!20}{-}&\cellcolor{yellow!20}{-}&\cellcolor{yellow!20}{-}&\cellcolor{orange!20}{24.95}&\cellcolor{orange!20}{24.84}&\cellcolor{orange!20}{25.76}&\cellcolor{orange!20}{24.62}&\cellcolor{orange!20}{24.67}&\cellcolor{SkyBlue!40}25.24&\cellcolor{SkyBlue!40}24.63&\cellcolor{SkyBlue!40}24.98&\cellcolor{SkyBlue!20}25.57&\cellcolor{green!20}24.60&\cellcolor{green!20}25.64\\
    \hline
    \rowcolor{gray!20}
    \multicolumn{1}{r}{\textcolor{black}{\textit{Proprietary Models}}} &\multicolumn{15}{r}{} \\
    \hline
    \textbf{Gemini-2.5-pro} & \cellcolor{red!20}57.13 & \cellcolor{red!20}72.78 & \cellcolor{yellow!20}48.26 & \cellcolor{yellow!20}38.66 & \cellcolor{orange!20}69.39 & \cellcolor{red!20}62.24 & \cellcolor{red!20}78.79 & \cellcolor{red!20}62.24 & \cellcolor{red!20}91.00 & \cellcolor{red!20}73.20 & \cellcolor{red!20}94.90 & \cellcolor{red!20}86.46 & \cellcolor{red!20}80.81 & \cellcolor{red!20}73.33 & \cellcolor{red!20}74.23\\
    \hline
    \textbf{GPT-4o-mini} & \cellcolor{yellow!20}31.49 & \cellcolor{yellow!20}54.17 & \cellcolor{yellow!20}28.26 & \cellcolor{yellow!20}24.95 & \cellcolor{orange!20}73.47 & \cellcolor{orange!20}49.48 & \cellcolor{orange!20}26.53 & \cellcolor{orange!20}34.69 & \cellcolor{orange!20}20.00 & \cellcolor{SkyBlue!40}31.96 & \cellcolor{SkyBlue!40}9.18 & \cellcolor{SkyBlue!40}23.96 & \cellcolor{SkyBlue!20}33.33 & \cellcolor{green!20}0.00 & \cellcolor{green!20}31.96\\
    \hline
    \textbf{Claude-sonnet-4.5} & \cellcolor{yellow!20}41.59 & \cellcolor{yellow!20}70.51 & \cellcolor{yellow!20}40.87 & \cellcolor{yellow!20}41.67 & \cellcolor{orange!20}21.79 & \cellcolor{orange!20}13.58 & \cellcolor{orange!20}32.10 & \cellcolor{orange!20}33.33 & \cellcolor{orange!20}20.99 & \cellcolor{SkyBlue!40}36.25 & \cellcolor{SkyBlue!40}30.00 & \cellcolor{SkyBlue!40}26.58 & \cellcolor{SkyBlue!20}22.50 & \cellcolor{green!20}20.00 & \cellcolor{green!20}28.75\\
    \hline
    \textbf{GPT-5-chat-latest} & \cellcolor{yellow!20}38.33 & \cellcolor{yellow!20}72.15 & \cellcolor{yellow!20}42.61 & \cellcolor{yellow!20}36.48 & \cellcolor{orange!20}5.32 & \cellcolor{orange!20}3.19 & \cellcolor{orange!20}29.47 & \cellcolor{orange!20}17.02 & \cellcolor{orange!20}19.79 & \cellcolor{SkyBlue!40}16.13 & \cellcolor{SkyBlue!40}20.43 & \cellcolor{SkyBlue!40}10.87 & \cellcolor{SkyBlue!20}20.00 & \cellcolor{green!20}13.33 & \cellcolor{green!20}25.81\\
    \hline
    \rowcolor{gray!20}
    \multicolumn{1}{r}{\textcolor{black}{\textit{Open Source Models}}}&\multicolumn{15}{r}{} \\
    \hline
    \textbf{Cambrian-S-3B} & \cellcolor{yellow!20}46.69 & \cellcolor{yellow!20}30.00 & \cellcolor{yellow!20}24.24 & \cellcolor{yellow!20}19.86 & \cellcolor{orange!20}46.46 & \cellcolor{orange!20}42.73 & \cellcolor{orange!20}57.93 & \cellcolor{orange!20}40.27 & \cellcolor{orange!20}54.43 & \cellcolor{SkyBlue!40}50.88 & \cellcolor{SkyBlue!40}71.93 & \cellcolor{SkyBlue!40}73.21 & \cellcolor{SkyBlue!20}64.22 & \cellcolor{green!20}72.41 & \cellcolor{green!20}61.16\\
    \hline
    \textbf{VST-3B-RL} & \cellcolor{yellow!20}32.24 & \cellcolor{yellow!20}18.08 & \cellcolor{yellow!20}7.76 & \cellcolor{yellow!20}23.03 & \cellcolor{orange!20}53.54 & \cellcolor{orange!20}46.70 & \cellcolor{orange!20}54.63 & \cellcolor{orange!20}40.71 & \cellcolor{orange!20}53.59 & \cellcolor{SkyBlue!40}47.35 & \cellcolor{SkyBlue!40}61.40 & \cellcolor{SkyBlue!40}73.21 & \cellcolor{SkyBlue!20}69.83 & \cellcolor{green!20}73.28 & \cellcolor{green!20}56.70\\
    \hline
    \textbf{Qwen3-VL-235B-A22B-Instruct} & \cellcolor{yellow!20}43.76 & \cellcolor{yellow!20}70.10 & \cellcolor{yellow!20}40.43 & \cellcolor{yellow!20}34.33 & \cellcolor{orange!20}68.37 & \cellcolor{orange!20}24.49 & \cellcolor{orange!20}30.30 & \cellcolor{orange!20}32.99 & \cellcolor{orange!20}31.00 & \cellcolor{SkyBlue!40}36.08 & \cellcolor{SkyBlue!40}25.51 & \cellcolor{SkyBlue!40}37.50 & \cellcolor{SkyBlue!20}33.33 & \cellcolor{green!20}26.67 & \cellcolor{green!20}43.30\\
    \hline
    \textbf{Qwen3-VL-235B-A22B-Thinking} & \cellcolor{yellow!20}45.36 & \cellcolor{yellow!20}71.03 & \cellcolor{yellow!20}43.18 & \cellcolor{yellow!20}27.73 & \cellcolor{orange!20}45.92 & \cellcolor{orange!20}31.58 & \cellcolor{orange!20}31.63 & \cellcolor{orange!20}20.41 & \cellcolor{orange!20}37.00 & \cellcolor{SkyBlue!40}44.33 & \cellcolor{SkyBlue!40}17.35 & \cellcolor{SkyBlue!40}27.08 & \cellcolor{SkyBlue!20}28.28 & \cellcolor{green!20}40.00 & \cellcolor{green!20}45.36\\
    \hline
    \textbf{Qwen2.5-VL-72B-Instruct} & \cellcolor{yellow!20}26.14 & \cellcolor{yellow!20}57.94 & \cellcolor{yellow!20}43.48 & \cellcolor{yellow!20}30.21 & \cellcolor{orange!20}37.76 & \cellcolor{orange!20}13.27 & \cellcolor{orange!20}27.27 & \cellcolor{orange!20}36.73 & \cellcolor{orange!20}30.00 & \cellcolor{SkyBlue!40}40.21 & \cellcolor{SkyBlue!40}16.33 & \cellcolor{SkyBlue!40}43.75 & \cellcolor{SkyBlue!20}36.36 & \cellcolor{green!20}6.67 & \cellcolor{green!20}37.11\\
    \hline
    \textbf{Qwen3-VL-30B-A3B-Instruct} & \cellcolor{yellow!20}48.02 & \cellcolor{yellow!20}59.28 & \cellcolor{red!20}63.04 & \cellcolor{yellow!20}34.85 & \cellcolor{red!20}74.49 & \cellcolor{orange!20}58.16 & \cellcolor{orange!20}30.30 & \cellcolor{orange!20}25.51 & \cellcolor{orange!20}21.00 & \cellcolor{SkyBlue!40}35.05 & \cellcolor{SkyBlue!40}6.12 & \cellcolor{SkyBlue!40}22.92 & \cellcolor{SkyBlue!20}29.29 & \cellcolor{green!20}0.00 & \cellcolor{green!20}38.14\\
    \hline
    \textbf{Qwen3-VL-30B-A3B-Thinking} & \cellcolor{yellow!20}48.79 & \cellcolor{yellow!20}63.40 & \cellcolor{yellow!20}27.83 & \cellcolor{yellow!20}30.72 & \cellcolor{orange!20}71.43 & \cellcolor{orange!20}45.36 & \cellcolor{orange!20}31.31 & \cellcolor{orange!20}27.55 & \cellcolor{orange!20}32.00 & \cellcolor{SkyBlue!40}35.05 & \cellcolor{SkyBlue!40}11.22 & \cellcolor{SkyBlue!40}32.29 & \cellcolor{SkyBlue!20}26.26 & \cellcolor{green!20}13.33 & \cellcolor{green!20}30.93\\
    \hline
    \textbf{GLM-4V-Plus-0111} & \cellcolor{yellow!20}45.70 & \cellcolor{yellow!20}65.98 & \cellcolor{yellow!20}40.45 & \cellcolor{yellow!20}39.90 & \cellcolor{orange!20}60.20 & \cellcolor{orange!20}55.67 & \cellcolor{orange!20}27.84 & \cellcolor{orange!20}36.73 & \cellcolor{orange!20}24.24 & \cellcolor{SkyBlue!40}34.02 & \cellcolor{SkyBlue!40}14.29 & \cellcolor{SkyBlue!40}22.92 & \cellcolor{SkyBlue!20}36.73 & \cellcolor{green!20}13.33 & \cellcolor{green!20}18.56\\
    \hline
    \textbf{ERNIE-4.5-Turbo-VL-32K} & \cellcolor{yellow!20}44.06 & \cellcolor{yellow!20}64.85 & \cellcolor{yellow!20}24.78 & \cellcolor{yellow!20}28.45 & \cellcolor{orange!20}72.45 & \cellcolor{orange!20}60.20 & \cellcolor{orange!20}25.25 & \cellcolor{orange!20}27.55 & \cellcolor{orange!20}23.00 & \cellcolor{SkyBlue!40}24.74 & \cellcolor{SkyBlue!40}10.20 & \cellcolor{SkyBlue!40}27.08 & \cellcolor{SkyBlue!20}30.30 & \cellcolor{green!20}0.00 & \cellcolor{green!20}21.65\\
    \hline
    \textbf{GLM-4.5V} & \cellcolor{yellow!20}48.53 & \cellcolor{yellow!20}66.60 & \cellcolor{yellow!20}23.04 & \cellcolor{red!20}41.75 & \cellcolor{red!20}74.49 & \cellcolor{orange!20}28.87 & \cellcolor{orange!20}32.32 & \cellcolor{orange!20}29.59 & \cellcolor{orange!20}26.00 & \cellcolor{SkyBlue!40}33.68 & \cellcolor{SkyBlue!40}8.16 & \cellcolor{SkyBlue!40}33.68 & \cellcolor{SkyBlue!20}28.12 & \cellcolor{green!20}26.67 & \cellcolor{green!20}15.46\\
    \hline
    \textbf{Qwen2.5-VL-7B-Instruct} & \cellcolor{yellow!20}19.21 & \cellcolor{yellow!20}28.04 & \cellcolor{yellow!20}17.39 & \cellcolor{yellow!20}12.37 & \cellcolor{orange!20}18.37 & \cellcolor{orange!20}11.34 & \cellcolor{orange!20}11.11 & \cellcolor{orange!20}16.33 & \cellcolor{orange!20}20.20 & \cellcolor{SkyBlue!40}15.05 & \cellcolor{SkyBlue!40}14.74 & \cellcolor{SkyBlue!40}23.96 & \cellcolor{SkyBlue!20}18.95 & \cellcolor{green!20}33.33 & \cellcolor{green!20}13.54\\
    \bottomrule
    \end{tabular}
    }
\end{table*}

\begin{table}[h]
\centering
\caption{Evaluation result on few-shot setting with $\Delta$ change. \textbf{Bold} indicates positive delta over 10 points on few-shot setting.}
\label{tab:optimized-few-shot}
\small
\resizebox{\textwidth}{!}{
\begin{tabular}{llllllll}
\toprule
\textbf{Model \& Setting} & \textbf{Score} & \textbf{Ave.} & \textbf{Obs.} & \textbf{Topo.} & \textbf{Symb.} & \textbf{Caus.} & \textbf{Plan.} \\ \midrule
\textbf{Gemini-2.5-pro} & & & & & & & \\
Default & 75.79 & 71.86 & 55.63 & 72.82 & 84.88 & 80.81 & 74.11 \\
Few-shot & 68.12 {\color{red}\scriptsize(-7.7)} & 67.43 {\color{red}\scriptsize(-4.4)} & 56.15 {\color{blue}\scriptsize(+0.5)} & 66.73 {\color{red}\scriptsize(-6.1)} & 80.00 {\color{red}\scriptsize(-4.9)} & 66.67 {\color{red}\scriptsize(-14.1)} & 66.96 {\color{red}\scriptsize(-7.2)} \\ 

\midrule

\textbf{GPT-5-chat-latest} & & & & & & & \\
Default & 22.45 & 23.93 & 48.48 & 15.01 & 15.83 & 20.00 & 24.07 \\
Few-shot & \textbf{61.08} {\color{blue}\scriptsize(+38.6)} & \textbf{59.03} {\color{blue}\scriptsize(+35.1)} & 50.16 {\color{blue}\scriptsize(+1.7)} & \textbf{55.98} {\color{blue}\scriptsize(+41.0)} & \textbf{72.41} {\color{blue}\scriptsize(+56.6)} & \textbf{56.57} {\color{blue}\scriptsize(+36.6)} & \textbf{64.29} {\color{blue}\scriptsize(+40.2)} \\ 

\midrule

\textbf{Qwen3-VL-235B-A22B-Inst} & & & & & & & \\
Default & 37.79 & 38.84 & 48.68 & 37.40 & 32.99 & 33.33 & 41.07 \\
Few-shot & \textbf{63.20} {\color{blue}\scriptsize(+25.4)} & \textbf{61.26} {\color{blue}\scriptsize(+22.4)} & 47.29 {\color{red}\scriptsize(-1.4)} & \textbf{63.29} {\color{blue}\scriptsize(+25.9)} & \textbf{72.76} {\color{blue}\scriptsize(+39.8)} & \textbf{53.54} {\color{blue}\scriptsize(+20.2)} & \textbf{70.54} {\color{blue}\scriptsize(+29.5)} \\ 

\midrule

\textbf{VST-3b-RL} & & & & & & & \\
Default & 58.29 & 48.35 & 22.71 & 49.87 & 60.62 & 69.83 & 62.35 \\
Few-shot & 35.56 {\color{red}\scriptsize(-22.73)} & 34.44 {\color{red}\scriptsize(-13.91)} & 7.26 {\color{red}\scriptsize(-15.45)} & 42.62 {\color{red}\scriptsize(-7.25)} & 50.69 {\color{red}\scriptsize(-9.93)} & 28.57 {\color{red}\scriptsize(-41.26)} & 37.84 {\color{red}\scriptsize(-24.51)} \\ 

\midrule

\textbf{Cambrian-S-3B} & & & & & & & \\
Default & 59.12 & 50.90 & 31.67 & 48.43 & 65.34 & 64.22 & 65.0 \\
Few-shot & 50.35 {\color{red}\scriptsize(-8.77)} & 48.49 {\color{red}\scriptsize(-2.41)} & 26.08 {\color{red}\scriptsize(-5.59)} & 52.87 {\color{blue}\scriptsize(+4.44)} & 62.15 {\color{red}\scriptsize(-3.19)} & 40.82 {\color{red}\scriptsize(-23.4)} & 57.66 {\color{red}\scriptsize(-7.34)} \\ 

\midrule

\textbf{Qwen2.5-VL-7B-I} & & & & & & & \\
Default & 17.45 & 17.84 & 19.69 & 15.48 & 17.96 & 18.95 & 16.22 \\
Few-shot & 24.85 {\color{blue}\scriptsize(+6.69)} & 24.41 {\color{blue}\scriptsize(+7.01)} & 12.55 {\color{red}\scriptsize(-7.14)} & 25.79 {\color{blue}\scriptsize(+10.21)} & 36.02 {\color{blue}\scriptsize(+18.06)} & 21.88 {\color{blue}\scriptsize(+2.93)} & 24.29 {\color{blue}\scriptsize(+8.07)} \\ 

\midrule

\textbf{Qwen3VL-30B-A3B-I} & & & & & & & \\
Default & 32.83 & 37.07 & 48.52 & 41.78 & 21.31 & 29.29 & 33.04 \\
Few-shot & 54.19 {\color{blue}\scriptsize(+21.36)} & 52.39 {\color{blue}\scriptsize(+15.32)} & 44.24 {\color{red}\scriptsize(-4.28)} & 51.02 {\color{blue}\scriptsize(+2.5)} & 60.42 {\color{blue}\scriptsize(+39.11)} & 46.94 {\color{blue}\scriptsize(+17.65)} & 61.26 {\color{blue}\scriptsize(+28.22)} \\

\bottomrule
\end{tabular}
}
\end{table}

\section{Evaluation on SpatialBench}
\label{sec:eva}
\subsection{Setup}
\noindent\textbf{Models.} We evaluate a diverse set of MLLMs to comprehensively assess their spatial cognitive abilities. Our benchmark covers both proprietary and open-source models with support for video understanding. For proprietary models, we include leading systems such as Gemini \cite{gemini2023gemini}, GPT \cite{hurst2024gpt,openai2025gpt5}, and Claude-Sonnet \cite{anthropic2025claudesonnet45}. For open-source models, we evaluate representative families including Qwen \cite{bai2023qwen}, GLM \cite{glm2024chatglm}, ERNIE \cite{zhang2019ernie}, and spatial-centric models VST \cite{yang2025vst} and Cambrian-S \cite{yang2025cambrian} encompassing variants across a broad range of architectures, parameter sizes (7B–235B), and training paradigms, enabling a systematic comparison of how different modeling strategies influence spatial cognition performance in SpatialBench.

\noindent\textbf{Metrics.} Following \cite{yang2025thinking}, we adopt evaluation metrics tailored to the answer type. Specifically, tasks in SpatialBench are categorized into Multiple-Choice Answer (MCA) and Numerical Answer (NA) formats. For MCA tasks, we employ accuracy (ACC) as the primary evaluation metric \cite{fu2025videomme,hendrycks2021measuring}, which measures the proportion of exactly matched answers between the model’s predictions and the ground-truth labels. While for NA tasks, where answers involve continuous numerical values, we adopt the mean relative accuracy (MRA) metric \cite{yang2025thinking,lin2014microsoft}, defined as:
\begin{equation}
    \text{MRA}=\frac{1}{10}\sum_{\theta\in\Omega}\mathbb{I}(\frac{|y'-y|}{y}<1-\theta),
\end{equation}
where $y'$ and $y$ are the model's prediction and ground truth, respectively, and $\theta$ is the confidence threshold from a thresholds set $\Omega=\{0.5,0.55,...,0.95\}$.

\noindent\textbf{Overall Score.} We propose a high-level capability–oriented overall score to integrate performance across the five hierarchical cognitive levels. To emphasize higher-level reasoning while preserving balance, we assign adaptive weights to each level:
\begin{equation}
    C_i = F_iS_i, \quad F_i = \alpha D_i + 0.1(1-\alpha)E_i,\ i=1,...,5,
\end{equation}
where $S_i$ is the standard deviation of model scores, $D_i$ is the category’s question proportion, and $\alpha$ controls the trade-off between baseline and adaptive weights and $E_i$ is non-negative variables to be optimized with $\sum_{i=1}^5 E_i=10$. Under the monotonicity constraint $C_{i+1}>C_i$ the target is to minimize
\begin{equation}
    \text{Var}(C_{i+1}-C_i) - k\sum_{j=1}^5 C_j^2,
\end{equation}
yielding complexity-aware overall scores. $\alpha \ \text{and}\  k$ will be fixed manually. Finally, the overall score is computed as $\sum_{i=1}^{5} C_iM_i$, where $M_i$ is the average rating on each level.

\subsection{Benchmarking MLLM Performance}
Table \ref{tab:result-com} and Table \ref{tab:result} presents the model performance on SpatialBench across five hierarchical levels. Gemini-2.5-pro occupy the top of the ranking and show substantially higher overall scores than any other models. This gap is most pronounced on high-level tasks such as symbolic reasoning, causality, and planning. Nevertheless, several open-source series (e.g., Qwen variants) reach competitive performance on lower and mid levels, indicating that open models can achieve strong perceptual and relational understanding with carefully designed training paradigms. For open-source models, a clear correlation appears between model scale and average performance: larger models generally achieve higher overall scores, showing that scale remains an important factor. However, size alone does not guarantee better performance. Within the same model family, different versions (such as instruction-tuned vs. thinking mode) can show clear performance gaps. This suggests that architectural design and the integration strategy for visual and linguistic information plays a crucial role in spatial cognitive ability. Additionally, spatial-centric models such as VST and Cambrian-S rank among the top open-source models despite their relatively small parameter scales. This suggests that spatial-intelligence-oriented reinforcement learning (RL, as in VST) and instruction tuning (IT, as in Cambrian-S) can substantially improve performance, particularly when compared with their backbone model, Qwen2.5-VL.

It can be observed that a cross the board, observation and topology tasks are relatively easier for MLLMs. Most models achieve substantially higher scores on tasks such as object counting, size and distance estimation, and simple topological queries. By contrast, higher-level abilities remain challenging for models, including symbolic reasoning, causality, and visual planning. On these tasks, average performance drops noticeably and variance between models increases. This pattern suggests that while current MLLMs can reliably extract visual evidence and reason about basic relations, they struggle to (a) convert perceptual inputs into robust symbolic rules, (b) infer causal or dynamic consequences accurately, and (c) generate multi-step and convincing plans for a given objective.

\begin{figure}
    \centering
    \includegraphics[width=.98\linewidth]{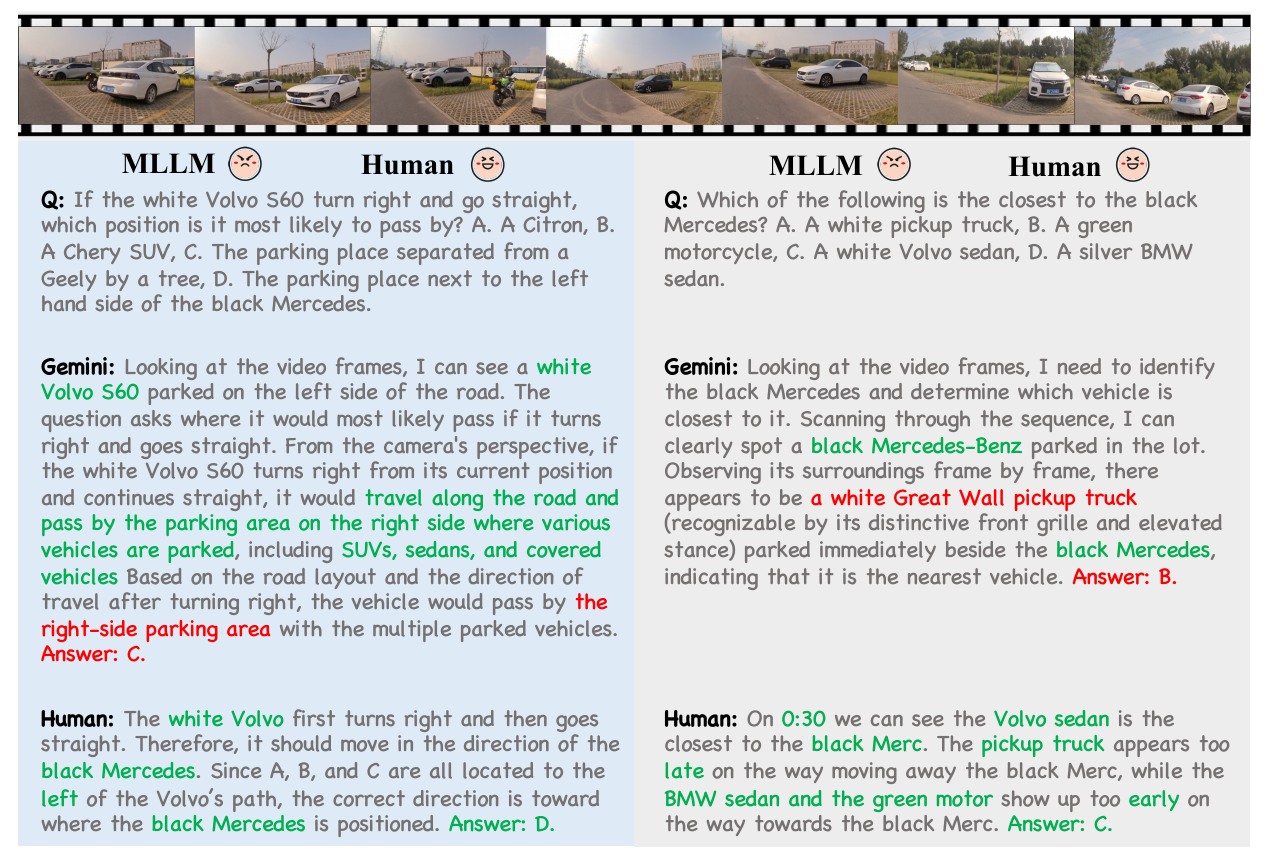}
    \caption{The differences of thinking processes between MLLM and human.}
    \label{fig:example}
\end{figure}
\vspace{-0.6cm}
\subsection{One-Shot Evaluation}
In-context learning is a critical capability for MLLMs, reflecting their ability to leverage minimal examples for reasoning. To evaluate this, we perform a one-shot assessment: for each task, a single annotated example comprising a QA pair, reasoning explanation, and key frames is provided. Models are then asked to answer a test question from the same task, enabling examinations on how effectively they can generalize spatial reasoning from limited guidance. 

We select several representative models from the one-shot evaluation, including two proprietary models: Gemini-2.5-pro (the best-performing) and GPT-5-chat-latest (the weakest), as well as a strong open-source model Qwen3-VL-235B-A22B-Instruct. Gemini-2.5-pro shows a decline in performance, primarily across the intermediate cognitive levels. In contrast, GPT-5-chat-latest and Qwen3-VL demonstrate substantial improvements. Although these models still trail Gemini in absolute scores, their performance under prompting approaches that of the previously strongest model. This result indicates that even lower-performing models can rapidly enhance specific spatial reasoning abilities when given minimal in-context guidance, highlighting the potential of one-shot prompting to boost MLLMs’ higher-level cognitive capabilities. The results suggest that some powerful models, such as Gemini 2.5 Pro, can reason efficiently and accurately from context on their own, performing better than when given human-guided prompts. This indicates that their built-in contextual understanding is particularly well-suited for spatial intelligence tasks. By contrast, GPT‑5 appears comparatively weaker in intrinsic spatial logic, yet achieves more advanced spatial intelligence by leveraging the GPT series’ steadily improving linguistic capabilities \cite{achiam2023gpt}. Similarly, Qwen3-VL, which demonstrates strong text-based reasoning skills \cite{qwen3vl-hf}, benefits from one-shot prompting in a way comparable to GPT‑5, highlighting the role of linguistic reasoning in enhancing spatial cognition under minimal guidance.


\subsection{Benchmarking Human Performance}
To quantify the gap between MLLMs and human-level intelligence across cognitive dimensions, we conduct a human benchmark experiment. In this experiment, 33 human participants are presented with SpatialBench, and they provide answers directly without guidance, allowing us to measure their unaided spatial reasoning and planning abilities. The results in Table \ref{tab:result-com} and Table \ref{tab:result} indicate that human achieve near-perfect performance across nearly all tasks with an overall score of 96.40. Humans perform particularly well on higher-level tasks: symbolic reasoning, causality, and planning, and all achieve essentially 100\% accuracy, while even lower-level observation and topology tasks maintain strong performance. Across all levels, humans exhibit not only higher absolute accuracy but also more consistent performance, reflecting robust generalization and contextual understanding that MLLMs have yet to fully achieve. These findings underscore the substantial gap that still exists between human and machine spatial intelligence. While MLLMs show promising abilities in extracting visual information and reasoning about simple topological relations, they remain far from matching human performance in tasks that require integrating high-level reasoning. This test thus provides a clear target for future model development, highlighting the need to enhance multi-step reasoning and context-sensitive planning in MLLMs.

Fig. \ref{fig:example} shows an example of the differences between Gemini and human. The human focuses on the key directional cue: the turning path of the white Volvo, and quickly eliminates irrelevant options based on spatial orientation, demonstrating strong goal-directed and spatially grounded reasoning. In contrast, Gemini describes the entire scene in a more exhaustive but unfocused manner, mentioning many vehicles and areas without identifying the crucial spatial relationship. This suggests that while MLLMs can recognize objects and describe scenes accurately, they often lack selective attention and directional understanding, leading them to infer by association rather than by reasoning about movement and geometry.

\subsection{Key Insights from Evaluation}
Our evaluation reveals several critical limitations in the spatial cognition of current MLLMs. First, models inherently struggle to reconstruct and maintain continuous spatial scenes from dynamic visual inputs. Poor-performing models frequently misinterpret relative directions or confuse sequential positions, while even advanced models make notable errors in continuous scene tracking and discerning subtle geometric relations. This highlights a fundamental vulnerability in capturing long-range spatial dependencies. Second, our findings empirically validate that spatial cognition in MLLMs is strictly hierarchical and tightly coupled: success in higher-order reasoning is fundamentally bottlenecked by accurate lower-level perception. Performance precipitously declines as cognitive complexity increases, with most models exhibiting cascading failures from the third level onward. To bridge these gaps, future work should leverage high-quality datasets structured with progressive cognitive tasks, incorporate explicit spatial representations, such as  scene graphs or 3D geometric priors, and adopt advanced training paradigms, including curriculum learning and agent-based interactive environments, to intrinsically promote multi-step reasoning and long-horizon planning. Furthermore, establishing closed-loop feedback mechanisms, where high-level reasoning and planning errors are utilized to iteratively refine low-level perceptual modules, could significantly mitigate the risk of cascading spatial hallucinations. 
\section{Conclusion}
\label{sec:con}
In this work, we introduce SpatialBench, a comprehensive benchmark built upon a five-level hierarchical spatial cognition framework that progressively evaluates MLLMs from low-level observation to high-level planning. This layered design reflects the cognitive progression from perception to decision-making, enabling a more interpretable and fine-grained diagnosis of multimodal spatial intelligence. Experimental results show that while modern MLLMs demonstrate strong perception and relational reasoning, their abilities in symbolic abstraction, causal inference, and spatial planning remain limited. SpatialBench establishes a principled foundation for hierarchical evaluation and future development of spatially grounded intelligence in MLLMs.


%
%
\bibliographystyle{splncs04}
\bibliography{main}

\appendix
\section{Mathematical Modeling of Overall Score}
To provide a unified measure of a model’s spatial cognitive competence, we introduce an overall score that integrates performance across all five hierarchical cognitive levels. The goal of this metric is not merely to average accuracy, but to construct a complexity-aware evaluation that reflects the progressively demanding nature of spatial cognition. Lower levels mainly involve perceptual and geometric understanding, while higher levels require abstract reasoning, causal inference, and planning. A meaningful overall score must therefore (1) preserve the relative importance of each cognitive level, (2) emphasize higher-level reasoning without overwhelming lower-level contributions, and (3) maintain fairness across categories with different question counts and variances.

To achieve these goals, we design a weighting mechanism that adaptively adjusts each level’s contribution based on its intrinsic difficulty and score distribution. Instead of manually assigning fixed weights, we formulate an optimization-driven approach that learns monotonic, complexity-aligned weights while controlling the imbalance between levels. This results in an overall metric that is interpretable, robust to distributional differences across task categories, and sensitive to a model’s true cognitive progression rather than raw accuracy alone.

To formalize the construction of our overall score, we assign adaptive weights to the five cognitive levels ($i=1,2,3,4,5$ corresponding to Observation through Planning). Our goal is to encourage higher performance on more complex cognitive abilities while preserving a balanced contribution across levels. 

We begin by computing, for each level, the empirical standard deviation
\begin{equation}
   S_i,\ i=1,2 ... 5,
\end{equation}
which reflects the intrinsic difficulty and discriminative range of that category. This quantity is paired with the category’s question proportion (i.e., its initial weight)
\begin{equation}
   D_i,\ i=1,2 ...5,
\end{equation}
serving as the baseline weighting factor. To introduce controlled adaptivity, we adjust these baseline weights through
\begin{equation}
   F_i=\alpha D_i+0.1(1-\alpha )E_i,
\end{equation}
where $\alpha$ is a manually chosen hyperparameter governing the balance between the baseline distribution $D_i$ and the optimized adjustment $E_i$. The resulting effective weight for each category is then
\begin{equation}
   C_i=F_iS_i,\ i=1,2 ... 5,
\end{equation}
which we require to increase monotonically with cognitive complexity.

To obtain a smooth hierarchy of difficulty, we further aim to make the increments between adjacent levels as uniform as possible. This leads to the following constrained nonlinear optimization problem:
\begin{equation}
   \begin{aligned}
       \min \  & f = \text{Var}\ ( C_{i+1} - C_i ) - k\sum_{j=1}^5 C_j^2, \quad i=1,2,3,4 \\
       \text{(P)} \quad &
       \begin{cases}
           C_{i+1} - C_i > 0,\ i=1,2,3,4\\
           \sum_{j=1}^5 E_j = 10\\
           E_i \ge 0,\ i=1,2 ...5
       \end{cases}
   \end{aligned},
\end{equation}
where the parameter $k$ expresses the preference for maintaining the original baseline separation prescribed by the standard deviations $S_i$. Solving Problem (P) yields a set of complexity-aware weights $C_i$, which are subsequently combined with the per-level average model ratings to produce the final overall score.

Mathematically, the original formulation introduces strict inequality constraints
\begin{equation}
    C_{j+1}-C_{j}>0, \ j=1,2,3,4,
\end{equation}
which renders the feasible region
\begin{equation}
    \{(E_i)\ |\ C_{j+1}-C_{j}>0\}
\end{equation}
an open set in Euclidean space. As a result, the constraint set is not compact and is incompatible with the standard form required by many numerical optimization solvers; specifically, it does not contain the closed and bounded simplex
\begin{equation}
    \{(E_i)\ |\ \sum_{j=1}^5 E_j=10,\ E_i\ge 0\}.
\end{equation}
Rather than transforming the problem into a fully standardized form, which would introduce unnecessary complications and yield no practical benefit, we relax the strict inequalities to non-strict ones and solve the following modified program:
\begin{equation}
   \begin{aligned}
       \min \  & f = \text{Var}\ ( C_{i+1} - C_i ) - k\sum_{j=1}^5 C_j^2, \quad i=1,2,3,4 \\
       (\text{P})' \quad &
       \begin{cases}
           C_{i+1} - C_i \ge 0,\ i=1,2,3,4\\
           \sum_{j=1}^5 E_j = 10\\
           E_i \ge 0,\ i=1,2 ...5
       \end{cases}
   \end{aligned}.
\end{equation}
The relaxation makes the feasible set closed and compatible with standard numerical solvers, while preserving the essential structural constraint that higher-level categories should not receive smaller weights than lower ones. After solving $\text{P})'$, we simply discard degenerate solutions in which
\begin{equation}
    C_i=C_{i+1},\ \exists\ i \in \{1,2,3,4\},
\end{equation}
since such solutions violate the intended strictly increasing hierarchy of cognitive complexity. This procedure is computationally effective and fully adequate for our application, as the optimization landscape naturally favors non-degenerate solutions when $k$ is chosen appropriately.

The new constraint set is now in standard form. Since both
\begin{equation}
\begin{aligned}
    & \{(E_i)\ |\ C_{j+1}-C_{j}>0, \ j=1,2,3,4\} \ \ \text{and}\\
    & \sum_{j=1}^5 E_j=10,\ E_i\ge 0\}
\end{aligned}
\end{equation}
are closed subsets of the Euclidean space, and their intersection (our new feasible region) is also closed.

Recall that $F_i=\alpha D_i+0.1(1-\alpha )E_i$ which is continuous in each $E_i$, Consequently, $C_i=F_iS_i$ is also continuous since $S_i$ is constant. Therefore,
\begin{equation}
    f_{sum}=\sum_{j=1}^5C_j^2
\end{equation}
is continuous as well. Let $\mathbf{E}=(E_1,E_2,...E_5)$, since each $C_i$ is a continuous function of $\mathbf{E}$, define
\begin{equation}
    g_i(\mathbf{E})=C_{i+1}(\mathbf{E})-C_i(\mathbf{E}), i=1,2,3,4,
\end{equation}
which is continuous in $\mathbf{E}$. Their linear combination
\begin{equation}
    \bar{g}(\mathbf{E})=\frac{1}{4}\sum_{i=1}^4 g_i(\mathbf{E})
\end{equation}
remains continuous. Thus,
\begin{equation}
    f_{var}=\text{Var}\ (g_i(\mathbf{E}))=\frac{1}{4}\sum_{i=1}^4(g_i(\mathbf{E})-\bar{g}(\mathbf{E}))^2
\end{equation}
is also continuous because it is constructed from addition and squaring of continuous functions. Hence,
\begin{equation}
    f=f_{var}-f_{sum}
\end{equation}
is a continuous function of $\mathbf{E}$. 

By the Extreme Value Theorem, any continuous function on a closed and bounded feasible region must attain its minimum. Therefore, problem $\text{P})'$ admits at least one global minimizer under the new constraint formulation.

To numerically solve the nonlinear programming problem, we employ the \verb|scipy.optimize.minimize| function in Python. We evaluate multiple combinations of $(\alpha,k)$, and observe that $(\alpha,k)=(0.4,0.01)$ yields a particularly favorable optimum. Under this setting, the optimizer returns
\begin{equation}
    (E_i)=(0,0,1.4911,3.8347,4.6742),
\end{equation}
achieving
\begin{equation}
    \text{Var}\ (C_{i+1}-C_i)=0.0264,
\end{equation}
which is substantially lower than the variance obtained when $\alpha=0$, where the optimal solution becomes
\begin{equation}
    (E_i)=\frac{2}{3}(1,2,3,4,5)
\end{equation}
with 
\begin{equation}
    \text{Var}\ (C_{i+1}-C_i)=0.0968.
\end{equation}
Moreover, the chosen parameter setting effectively resolves the initial undesirable ordering issue observed when $\alpha=1$, where the resulting sequence satisfies
\begin{equation}
    C_4<C_5<C_1<C_3<C_2,
\end{equation}
violating the monotonicity condition. In contrast, $(\alpha,k)=(0.4,0.01)$ produces a monotone and well-behaved solution consistent with our design constraints.

\section{Case Study}

\begin{figure*}
    \centering
    \includegraphics[width=1.\linewidth]{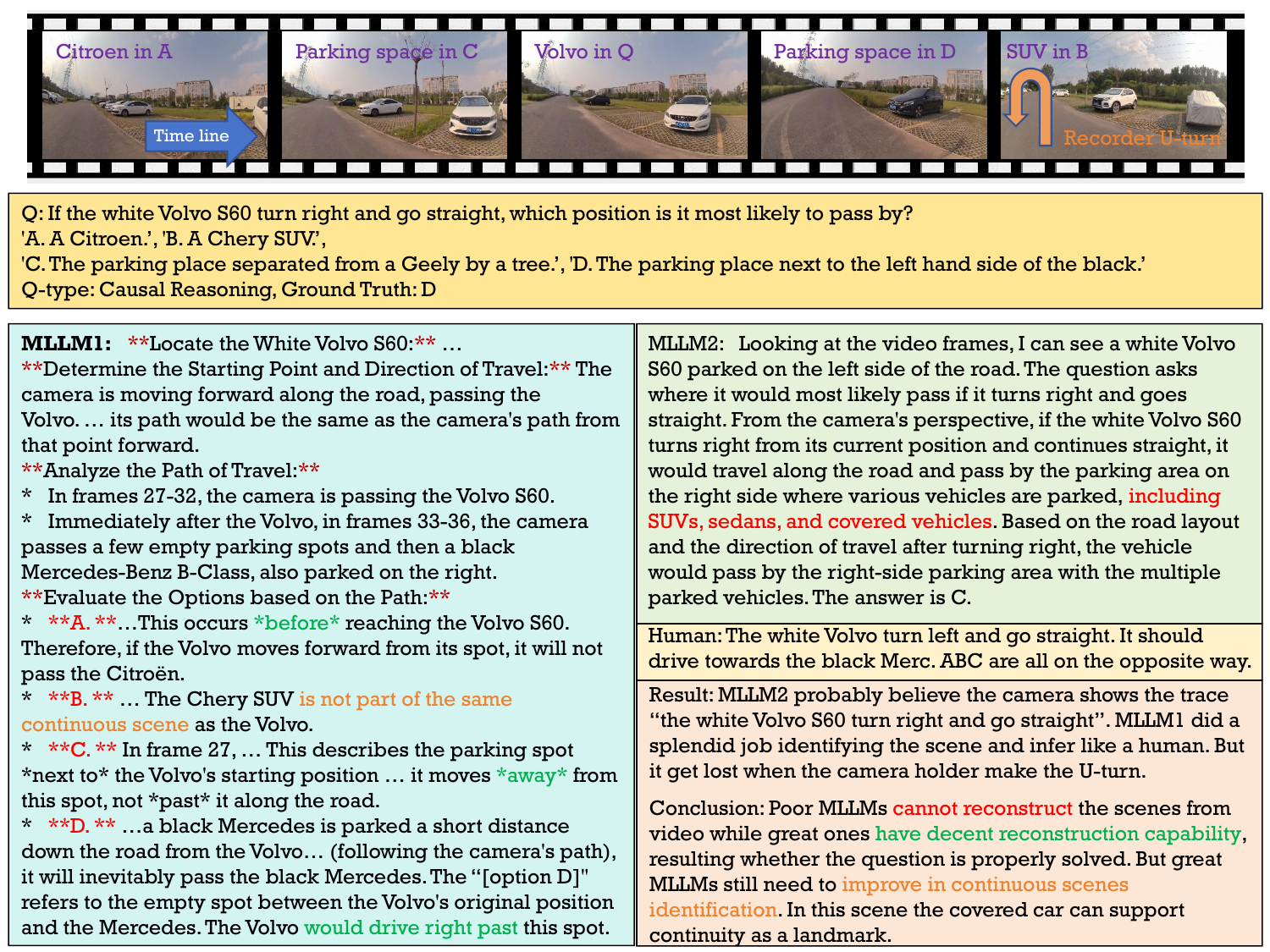}
    \caption{Case study on causal reasoning in dynamic scenes. This example shows that weaker MLLMs fail to construct a coherent scene, while stronger ones reconstruct the static layout but lose continuity when the camera changes direction.}
    \label{fig:conclusion}
\end{figure*}

\begin{figure*}
    \centering
    \includegraphics[width=1.\linewidth]{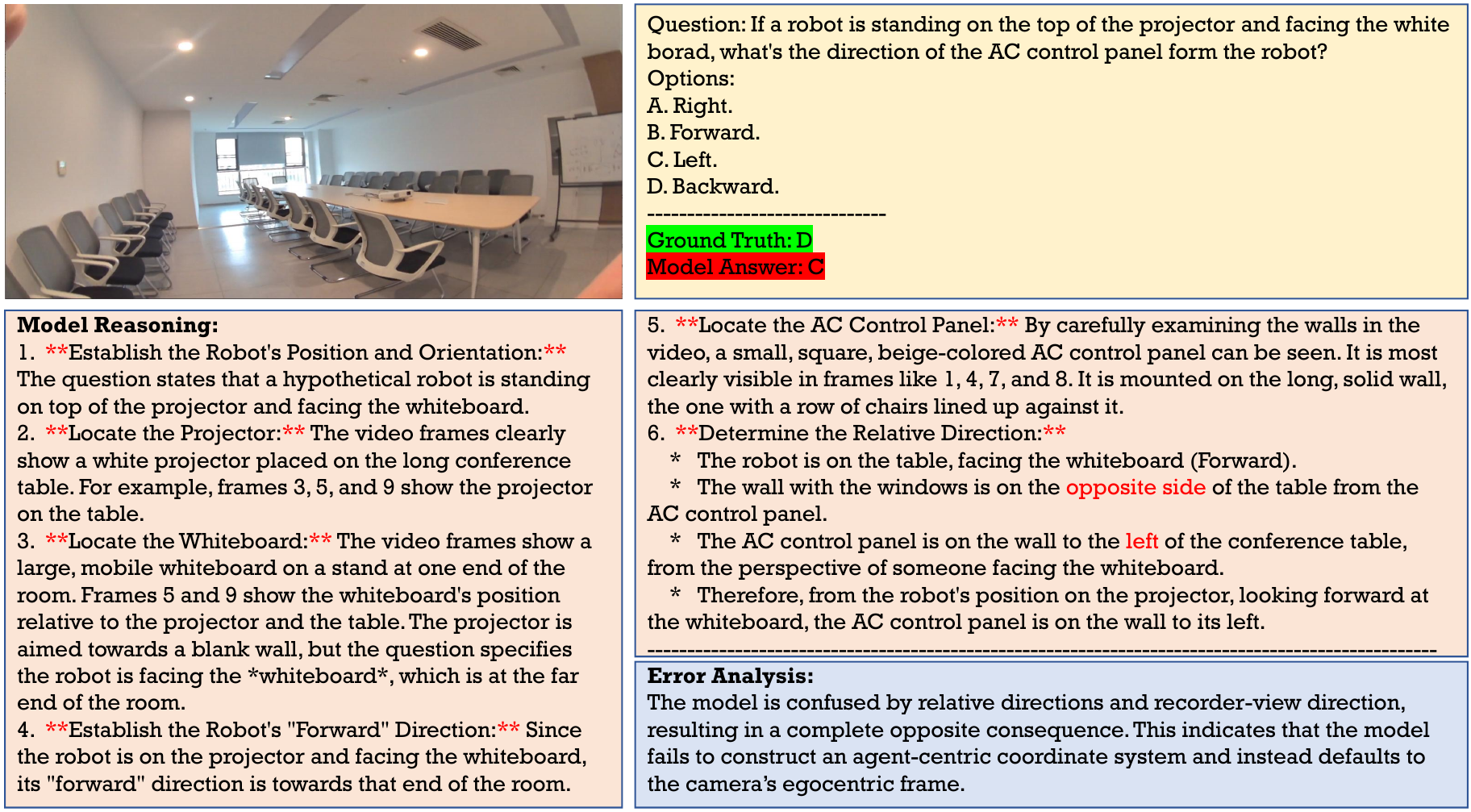}
    \caption{Failure case illustrating egocentric direction misinterpretation.
Although the camera is defined as facing the whiteboard, the model confuses recorder-view and agent-view directions, leading to the incorrect prediction (``Left'').}
    \label{fig:indoorcase}
\end{figure*}

To further illustrate the challenges of video-based spatial reasoning, we analyze a representative failure case from the causal reasoning category, and the results are shown in Figure \ref{fig:conclusion}. The question asks where a white Volvo S60 would most likely pass if it turns right and continues straight. While the ground truth is option D, the two evaluated MLLMs exhibit markedly different behaviors. MLLM1 demonstrates strong perceptual grounding: it accurately identifies the Volvo, reconstructs the forward scene layout, and reasons about nearby landmarks such as a black Mercedes and empty parking spots. However, its reasoning deteriorates once the camera performs a U-turn, causing the model to implicitly assume that the camera’s motion still reflects the Volvo’s hypothetical trajectory; this misalignment leads it to break the continuity of the reconstructed scene and misinterpret the spatial ordering. In contrast, MLLM2 fails much earlier and provides only a superficial description. It implicitly treats the camera’s viewing direction as the Volvo’s movement direction, confusing left/right relations and ultimately selecting an incorrect parking area. Human annotators, however, easily recognize that options A–C lie in the opposite direction of a right turn, and that the Volvo would naturally drive toward the black Mercedes, making D the only plausible answer. This case reveals a critical limitation: weaker models struggle to form any coherent scene representation, while stronger models can reconstruct static layouts but fail to maintain spatial consistency across continuous camera motion. Even when stable landmarks (e.g., the covered car) exist to support scene continuity, current models do not reliably leverage them. Overall, this example demonstrates that successful causal reasoning in dynamic scenes fundamentally depends on robust 3D scene reconstruction and continuity tracking—capabilities that remain insufficiently developed in existing MLLMs.

\section{Egocentric Reasoning Breakdown}
We further examine two representative failure cases involving egocentric direction reasoning in both indoor and outdoor environments. For the indoor case shown in Figure \ref{fig:indoorcase}, the task requires determining the relative direction of an AC control panel from the perspective of a robot standing on top of the projector and facing the whiteboard. While the ground truth is Backward, the model incorrectly predicts Left. A close inspection of the reasoning trace reveals that although the model correctly identifies the projector, the whiteboard, and the AC control panel within the room, it fails at the final spatial transformation: converting absolute room layout into the camera’s egocentric frame. The model implicitly adopts the recorder’s viewing direction as the reference frame, causing a systematic rotation of its inferred directions. As a result, the AC control panel which lies behind the camera when facing the whiteboard is erroneously mapped to its left. This case highlights a broader weakness: even when object localization is accurate, current MLLMs often conflate scene-centric, camera-centric, and agent-centric coordinate systems. Such confusion leads to consistent directional inversion or orthogonal errors, especially in indoor scenes where multiple frames provide shifting viewpoints. Strengthening explicit frame-of-reference reasoning remains essential for achieving reliable spatial understanding.

\begin{figure*}
    \centering
    \includegraphics[width=1.\linewidth]{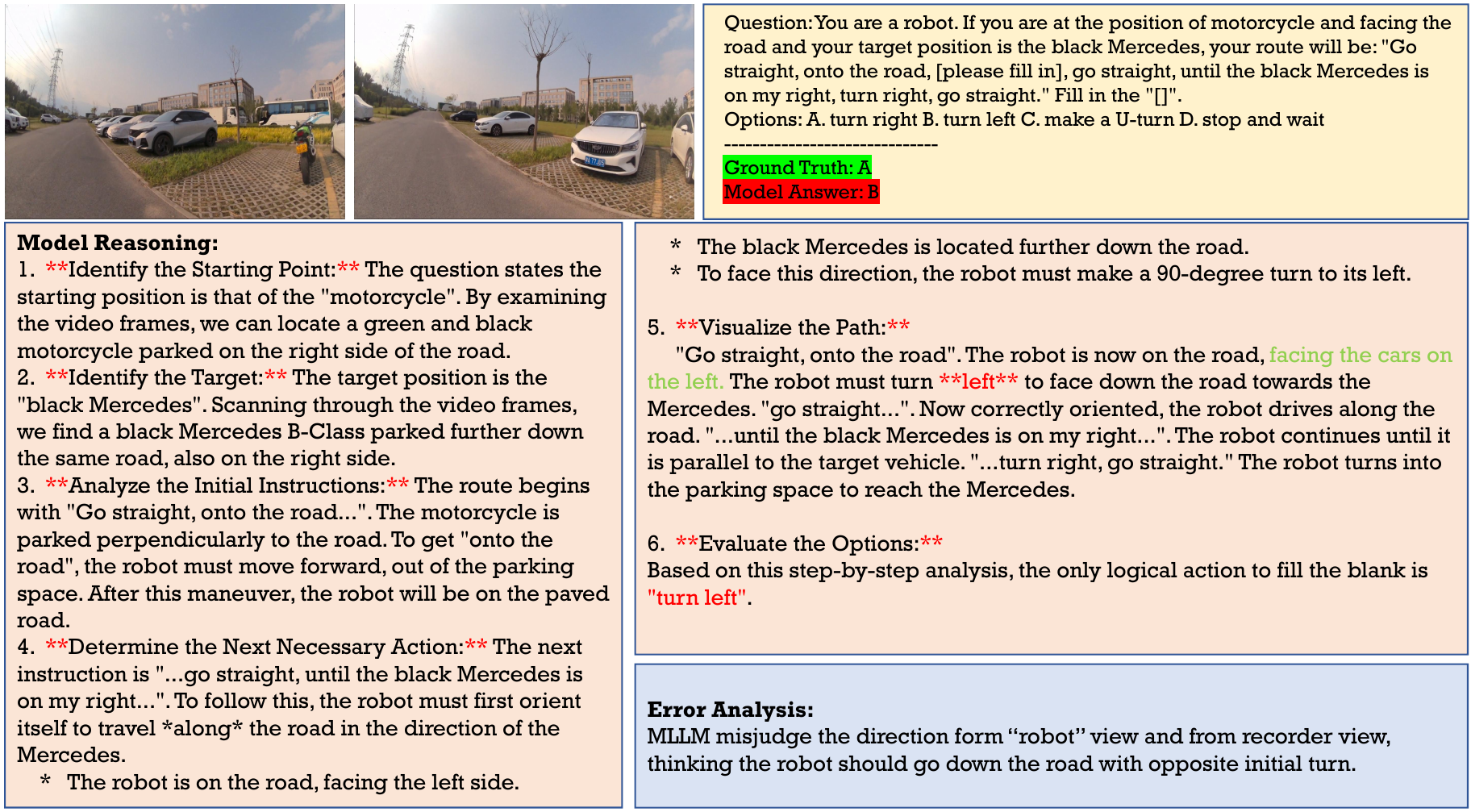}
    \caption{Failure case illustrating route-planning direction misinterpretation.
Although the robot’s forward orientation is clearly defined after moving onto the road, the model confuses recorder-view and agent-view directions, leading to the incorrect prediction (``Left'').}
    \label{fig:outdoorcase}
\end{figure*}

For the outdoor case in Figure \ref{fig:outdoorcase}, the model again fails to correctly align the robot’s egocentric perspective with the scene layout observed from the recorder’s view. Although the model successfully identifies the motorcycle and the black Mercedes, it misinterprets the orientation of the robot after moving ``onto the road.'' Because it interprets the forward direction based on the camera’s viewpoint rather than the robot’s own heading, the model incorrectly assumes that the robot should turn left to move down the road. This perspective confusion leads to a reversed decision in the route-planning step, causing the model to select “turn left” instead of the correct ``turn right.'' The error highlights the model’s difficulty in maintaining consistent egocentric orientation during multi-step spatial navigation.

\end{document}